\documentclass[runningheads]{llncs}
\usepackage[T1]{fontenc}
\usepackage{graphicx}
\usepackage{graphicx}
\usepackage{xspace,xcolor}
\usepackage{algorithm}
\usepackage[noend]{algpseudocode}

\usepackage{amsmath,amssymb,amsthm}
\usepackage{thmtools}
\usepackage{thm-restate}
\usepackage{mathtools}
\usepackage{tikz}
\usepackage{pgfplots}
\usepackage{hyperref}

\newcommand{\RLS}{\textsc{RLS}}

\newcommand{\LeadingOnes}{\textsc{LeadingOnes}}

\newcommand{\LO}{\textsc{LO}}

\newcommand{\new}[1]{\textcolor{black}{#1}}
\newcommand{\p}[1]{\textcolor{black}{#1}}
\newcommand{\xu}[1]{\textcolor{black}{#1}}
\newcommand{\xmq}[1]{\textcolor{black}{#1}}

\begin{document}
\title{On the Runtime Analysis of Reinforcement Learning Hyper-Heuristics
}
%
%
\author{Pietro S. Oliveto\thanks{Pietro S. Oliveto is the corresponding author.}
\orcidID{0000-0001-8164-6767} \and
Zhenyu Wang
\orcidID{0009-0007-6118-0737} \and
Peizhou Wu
\orcidID{0009-0000-3242-5423} \and
Mengqing Xu
\orcidID{0009-0007-8755-0871}
}
\authorrunning{P. S. Oliveto, Z. Wang, \xmq{P. Wu and M. Xu}}
\institute{Southern University of Science and Technology,\\ 518055 Shenzhen, China\\ 
\email{olivetop@sustech.edu.cn, wangzy20233@mail.sustech.edu.cn \\wupz2023@mail.sustech.edu.cn, xumq2025@mail.sustech.edu.cn}\\
}
\maketitle              
\begin{abstract}
%
Selection Hyper-heuristics (HHs) automate algorithmic design by selecting from a set of low-level heuristics which one to apply at each stage of the optimisation process. 
Several impressive results have been recently rigorously proven regarding the performance of selection hyper-heuristics (HHs) for standard benchmark functions.
However, the learning mechanisms employed by these HHs are considerably simplified compared to the machine learning techniques typically used in real world applications.
In this paper we analyse a Reinforcement Learning Hyper-heuristic (RLHH) from the literature.
The only previous result available proved that for a wide range of parameter settings, RLHH does not learn to select heuristics appropriately for the standard $\LeadingOnes$ benchmark function.
In this paper\xmq{,} we rigorously prove that with appropriate parameter values RLHH equipped with two random local search operators, $\RLS_1$ and $\RLS_2$ optimises the \xmq{$\LeadingOnes$} benchmark function in the best possible expected runtime achievable with the two operators up to lower order terms.
Experiments show that for realistic problem sizes it is faster than the Generalised Random Gradient HH which was previously proven to also have optimal expected runtime up to lower order terms.

\end{abstract}
\section{Introduction}
Hyper-heuristics (HHs) are automated algorithm design techniques for optimisation problems~\cite{Burke2013,PillayQuBook,DrakeEtAl2019}.
They work at a higher level of abstraction than randomised search heuristics by either generating offline a heuristic for the problem at hand
using heuristic components ({\it generation} HH) or online by selecting which heuristic to apply directly  at each stage of the optimisation process from a set of available low level 
heuristics ({\it selection} HH). The aim is to relieve the practitioner of the task of identifying which algorithm and related  parameter settings to apply to the given optimisation problem.
Furthermore, different heuristics, or even different parameter settings  may perform better at different stages of the optimisation process.

Numerous successful applications of HHs have been reported in several domains~\cite{Martins2017,Walker2016,Wu2016,Li2017,Kheiri2017,Ahmed2019,Kendall2013,Branke2016}. 
Although the field building a theoretical understanding of HHs is still in its infancy, several advances have been made recently regarding the theoretical foundations of selection HHs~\cite{LissovoiEtAl2023AIJ,MOW2025,DoerrIJCAI-25,DL2024,DDLS2023}. See~\cite{Oliveto2021} for a survey \xmq{of} earlier results. 


Selection HHs  typically consist of two components: 1) a heuristic methodology, also referred to as the {\it learning mechanism}, that selects which heuristic to be applied in the next step, and 2) a move-acceptance methodology to decide whether to accept the new solution. Concerning the move acceptance component it has been shown how simple move acceptance hyper-heuristics (MAHH) are very effective at escaping local optima. The MAHH switching with some probability $p$ between an elitist strict improvement acceptance rule and a non-elitist rule that accepts any generated solution  can optimise the standard \textsc{Cliff}$_d$ benchmark function in the best known expected runtime of $O(n^3/d^2 + n \log n)$~\cite{LissovoiEtAl2023AIJ}. A variant of the MAHH that switches between accepting strict improvements and only accepting worsening moves has been shown to optimise the \textsc{Jump}$_m$ function in an \xmq{impressive expected runtime of} $O((n^3/m) \log n)$ function evaluations~\cite{DoerrIJCAI-25}. 

Impressive results have also been shown when the performance of the HH depends on the learning mechanism used to determine the most appropriate low-level heuristic to be applied at each stage of the optimisation process. In particular, the Generalised Random Gradient (GRG) HH has been proven to optimise the standard \LeadingOnes~benchmark function in the optimal expected runtime (up to the leading constant) achievable by any unary unbiased search heuristic if equipped with a sufficient number of low-level Randomised Local Search (RLS) heuristics~\cite{LOW2020b}. Since the result depends on setting its {\it learning period} $\tau$ parameter appropriately, an Adaptive Random Gradient (ARG) was introduced later that also runs in optimal expected runtime, dispensing the user from having to identify an appropriate value for $\tau$~\cite{DOW2026}. These algorithms have also been shown to be effective for other classes of functions~\cite{LOW2020a}. Indeed, GRG has the best known expected runtime for the multimodal \textsc{TwoRates} benchmark function~\cite{MOW2025}. 

All the mentioned HHs use very simple learning mechanisms, the most sophisticated being the random gradient one that selects an operator uniformly at random and continues applying it so long as the operator succeeds at identifying an improving solution within the learning period $\tau$. These, however, are very different from the more sophisticated machine learning mechanisms typically employed by HHs in real-world applications \cite{Thabtah2008,Cowlingetal2003,Rattadilok2005,Burkeetal2003,N2003,McClymont2011,Bai2008,Ozcan2010,Blazewicz2011,Dowsland2007,Gibbs2010}. One exception is the work by Alanazi and Lehre~\cite{AL2016} which analysed a reinforcement learning HH (RLHH) from the HH literature~\cite{N2003}.
The RLHH assigns weights to each low level heuristic which are increased by a rewarding additive \xmq{term} $\alpha$ if the applied heuristic identifies an improvement and are decreased by a penalty \xmq{term} $\beta$ otherwise. Then the operator to be applied is decided at each step via roulette wheel selection.

The analysis of Alanazi and Lehre though only provided negative results by focusing on identifying parameter ranges for which the RLHH is inefficient.
In particular, they analysed the RLHH for the \LeadingOnes~benchmark function equipped with $m$ low-level heuristics among which only one has a positive probability of identifying improvements.
In this scenario they derived parameter settings for which  the RLHH provably cannot distinguish the effective operator from the rest, thus runs in the same expected runtime as an algorithm that at each step selects an operator uniformly at random (i.e., {\it simple random} HH~\cite{CowlingEtAl2000}).

In this paper we aim to \xmq{continue} building the theoretical foundations of HH using reinforcement learning mechanisms.
We prove that with appropriate parameter settings the RLHH equipped with two operators, $\RLS_1$ which flips one bit and $\RLS_2$ which flips two bits without replacement, runs in the best possible expected runtime achievable using the two operators (i.e. $\approx0.42329 n^2$) for \LeadingOnes. An experimental analysis reveals that RLHH is faster than the GRG HH for realistic problem sizes.

For space limitations we cannot include the proofs in this manuscript. 

\section{Preliminaries}
In this section, we formally define the Reinforcement Learning Hyper-Heuristic (RLHH) algorithm and the benchmark function used for the analysis. \p{We also introduce the mathematical tools that will be used in our runtime analysis.}
\subsection{Algorithm}
The Reinforcement Learning Hyper-heuristic (RLHH) framework analyzed in this work is a selection hyper-heuristic that dynamically manages a portfolio of $m$ low-level heuristics $H = \{h_1, \dots, h_m\}$. Unlike static selection mechanisms, RLHH \new{aims to} automatically learn the optimal heuristic to be applied at different stages of the optimization process by maintaining a weight $w_i$ for each heuristic. 

The pseudocode for the RLHH framework as considered in~\cite{AL2016} is given in Algorithm~\ref{alg:RLHH}.
\begin{algorithm}[t]
\caption{\textsc{Reinforcement Learning Hyper-heuristic (RLHH)} \cite{N2003,AL2016}}
\label{alg:RLHH}
    \begin{algorithmic}[1]
        \Procedure{RLHH}{$\alpha$,\,$\beta$,\,$w_{\min}$,\,$w_{\max}$, $w$}
            \State $s\gets \mathrm{Unif}(\mathcal{S}); w_{[m]}^{(0)}\gets w$
            \While{stopping conditions not satisfied}
                \State Pick $i\in[m]$, with probability $p_{i}^{(t)}=\frac{w_{i}^{(t)}}{\sum_{j=1}^{m}w_{j}^{(t)}}$
                \State $s^\prime\gets h_{i}(s)$
                \State $w_{i}^{(t+1)}\gets \begin{cases}
                    \min(w_{i}^{(t)}+\alpha,w_{\max}),&\mathrm{if}\  f(s^{\prime})>f(s)\\
                    \max(w_{i}^{(t)}-\beta,w_{\min}),&\mathrm{otherwise}
                \end{cases}$
                \If{$f(s^\prime)>f(s)$}
                    \State $s\gets s^\prime$
                \EndIf
            \EndWhile
            \State \Return $s$
        \EndProcedure
    \end{algorithmic}
\end{algorithm}
At each iteration, the algorithm selects a heuristic $h_i$ with a probability proportional to its current weight $p_i = w_i / \sum_{j=1}^m w_j$. If the applied heuristic generates a strict fitness improvement (i.e., $f(s') > f(s)$), it is rewarded by increasing its weight by an additive \xmq{term} $\alpha$, capped at a maximum value $w_{\max}$. Conversely, if the heuristic fails to strictly improve the fitness, it is penalized by subtracting a \xmq{term} $\beta$, bounded below by $w_{\min}$.
Other choices for the update rules (e.g., multiplicativex updates rather than additive) and for the selection rules (e.g., selecting the heuristic with highest weight i.e., {\it max choice}) have been applied in the literature. Here we follow the scheme analysed by Alanazi and Lehre described in Algorithm~\ref{alg:RLHH}~\cite{AL2016}.

In our analysis, we consider a portfolio of two standard mutation operators (i.e., $m=2$) commonly used in pseudo-Boolean optimization;
$h_1 = \RLS_1$ which flips exactly one bit chosen uniformly at random, and
$h_2 = \RLS_2$ which flips exactly two distinct bits chosen uniformly at random.

The only available runtime analysis of the RLHH has shown that if the parameters are set such that $\alpha$ is at most as large as $\beta$ (i.e. $\alpha \leq \beta$), then the HH
will not be able to distinguish between the low level heuristics, thus at each step chooses one uniformly at random~\cite{AL2016}.
For the hyper-heuristic to be effective, the reward \xmq{term} $\alpha$ should be able to counteract the weight decrease due to the expected number of failures before a success.

We configure the RLHH parameters as follows.
Reward \xmq{term} $\alpha = 1 + 1/\log n$, 
penalty \xmq{term} $\beta = 1/n$,
initial weight $w_1^{(0)} = w_2^{(0)} = \log n$,
upper bound $w_{\max} = \frac{n}{\log^{1+\gamma} n}$, where $\gamma \in (0, 1)$ is a strictly positive constant and 
lower bound $w_{\min} = 1$.
Considering that a basic randomised local search heuristic (i.e., $\RLS_1$) always has an improvement probability of  at least $1/n$ unless it is trapped on a local optima, thus succeeds in $n$ expected
steps, we set the penalty \xmq{term} to $\beta = 1/n$ and the reward \xmq{term} to $\alpha=1+ 1/\log n$ (slightly larger than $n$ times $\beta$). Thus, as long as the HH is not trapped on a local optima the weight of  $\RLS_1$ always increases in expectation. Yet, if some other low-level heuristic has a higher success probability, its weight will increase faster.
Since the operators are all initialised with the same weight value, we let the initial weight grow slightly with the problem size (i.e., $w_1^{(0)} = w_2^{(0)} = \log n$) so that the first operator to improve  does not gain a large weight advantage which may strongly bias (maybe incorrectly) the process in its favour.
An upper bound on the weights, $w_{\max}$, is required so that when the best operator for the current optimisation stage changes, the time for the respective weights to increase and decrease will not be too large. At the same time $w_{\max}$ should also not be too small, to avoid that the weight of $\RLS_1$ eventually catches up with that of the current best operator by reaching the upper bound due to its positive drift (by design). In this case it would not be possible to distinguish the weights because, once at the boundary,  the weight  of the best operator cannot increase further.
Our analyses will show that with a setting of $w_{\max} = \frac{n}{\log^{1+\gamma} n}$, $\gamma \in (0, 1)$ a constant, the RLHH optimises  \textsc{LeadingOnes}  in optimal expected runtime up to lower order terms.
 The lower bound on the weight can be simply set to any constant and we choose $w_{\min} = 1$.

\subsection{Benchmark Function}

We analyze the performance of the RLHH on the same standard pseudo-Boolean benchmark function \textsc{LeadingOnes} (LO) used in the previous work in the literature~\cite{AL2016}. 
The function is defined over the search space $\mathcal{S} = \{0, 1\}^n$ as
\[
\LO(s) = \sum_{i=1}^n \prod_{j=1}^i s_j.
\]
where $s_j$ is the value of the \xmq{$j$-th} bit in the bit string $s$.
The LO value of a bit-string is the number of consecutive leading 1-bits in the bit-string, thus 
the global optimum is the all-ones string $s^* = (1, \dots, 1)$ \xmq{since we are maximizing the function}. 

To successfully improve a candidate solution $s$ with current fitness $\LO(s) = i < n$, an algorithm must necessarily mutate the $(i+1)$-th bit (the first $0$-bit in the string). Any mutations applied to the leading prefix of $i$ ones will destroy the prefix and strictly decrease the fitness. Meanwhile, mutations applied to the trailing suffix (bits from position $i+2$ to $n$) are entirely invisible to the current fitness evaluation, acting as neutral moves.
The \xmq{unary} unbiased black box complexity of \textsc{LeadingOnes} is $\Theta(n^2)$~\cite{LehreWitt2012} and evolutionary algorithms using standard bit mutation match this performance.
The standard (1+1) EA (with mutation rate $\frac{1}{n}$) has an expected runtime of $\frac{e-1}{2}n^2-o(n^2)\approx0.85914n^2-o(n^2)$ \cite{BoettcherEtAl2010}.
By implementing a fitness-dependent mutation rate, the expected runtime of the (1+1) EA decreases to $\approx(1\pm o(1))0.68n^2$ \cite{BoettcherEtAl2010}.  
$\RLS_1$ has an expected runtime of $0.5n^2$ fitness function evaluations \cite{BuzdalovBuzdalova2015}.
The best expected runtime achievable by unbiased (1+1) black box algorithms is $\approx(1\pm o(1))0.388n^2$ \cite{DoerrWagner2018,Doerr2018Arxiv}.
The Generalised Random Gradient (GRG) and the Adaptive Random Gradient selection HHs with a sufficiently large constant sized low level heuristic set $|H|=k=O(1)$ have been shown to match this optimal performance~\cite{LOW2020b,DLOW2018,DOW2026}. If the heuristic set is limited to $\xmq{H}=\{\RLS_1,\RLS_2\}$, then the best possible expected runtime is $(1 + o(1)) \cdot 0.42329n^2$~\cite{LOW2020b}.
Here we show that RLHH matches this optimal performance.

We point out that other classes of randomised search heuristics with higher arity operators, including 
Estimation of Distribution Algorithms and Convex Search Algorithms, can run in $o(n^2)$ expected runtime \cite{AfshaniEtAl2013,FriedrichEtAl2016,MoraglioSudholt2017,DoerrKrejca2018}.


\subsection{Mathematical Tools}
To rigorously analyze the expected runtime and the stochastic dynamics of the weight adaptation mechanism of the RLHH algorithm, we rely on several fundamental tools from probability theory and stochastic processes. 


\begin{theorem}[Chernoff Bounds \textcolor{black}{\cite{M2017probability}}] \label{thm:chernoff}
Let $X = \sum_{i=1}^n X_i$ be the sum of independent Poisson trials such that $\mathbb{E}[X] = \mu$. Then for any $0 < \delta < 1$
\[
\Pr(X \le (1 - \delta)\mu) \le \exp\left(-\frac{\delta^2 \mu}{2}\right),
\]
and for any $\delta > 0$,
\[
\Pr(X \ge (1 + \delta)\mu) \le \left(\frac{e^\delta}{(1+\delta)^{1+\delta}}\right)^\mu.
\]
\end{theorem}


\begin{theorem}[Doob's Maximal Inequality~\cite{D2019probability}] \label{thm:doob}
Let $(Z_t)_{t \ge 0}$ be a stochastic process adapted to a filtration $(\mathcal{F}_t)_{t \ge 0}$. For any constant $\delta > 0$,
\begin{enumerate}
    \item[(i)] \textbf{Submartingale bound:} If $(Z_t)_{t \ge 0}$ is a non-negative submartingale, then for any finite time horizon $T \ge 0$, the probability that the maximum of the process reaches or exceeds $\delta$ is bounded by
    \[
    \Pr\left( \max_{0 \le t \le T} Z_t \ge \delta \right) \le \frac{\mathbb{E}[Z_T]}{\delta}.
    \]
    \item[(ii)] \textbf{Supermartingale bound:} If $(Z_t)_{t \ge 0}$ is a non-negative supermartingale, then the probability that the supremum of the process reaches or exceeds $\delta$ over infinite time is bounded by
    \[
    \Pr\left( \sup_{t \ge 0} Z_t \ge \delta \right) \le \frac{\mathbb{E}[Z_0]}{\delta}.
    \]
\end{enumerate}
\end{theorem}


\begin{theorem}[Freedman's Inequality\textcolor{black}{\cite{freedman1975}}, \cite{Tropp2011}] \label{thm:freedman}
Let $(Y_t)_{t \ge 1}$ be a martingale difference sequence with respect to a filtration $(\mathcal{F}_t)_{t \ge 0}$. Suppose there exists a constant $R$ such that $Y_t \le R$ almost surely for all $t \ge 1$. Let $M_k = \sum_{t=1}^k Y_t$ be the associated martingale, and let $V_k = \sum_{t=1}^k \operatorname{Var}(Y_t \mid \mathcal{F}_{t-1})$ be the predictable  quadratic variation. For any $\Delta > 0$ and $v > 0$,
\[
\Pr\left( \exists k \ge 1 : M_k \ge \Delta \text{ and } V_k \le v \right) \le \exp\left( - \frac{\Delta^2}{2(v + R\Delta/3)} \right).
\]
\end{theorem}


\begin{theorem}[Additive Drift Theorem\textcolor{black}{\cite{HeYao2001,LenglerBookChapter}}] \label{thm:additive-drift}
Let $(X_t)_{t \ge 0}$ be a stochastic process adapted to a filtration $(\mathcal{F}_t)_{t \ge 0}$ over a state space $\mathcal{S} \subseteq \mathbb{R}$. Let $T$ be the first hitting time of the region $(-\infty, a]$. If $X_0 \ge a$ and there exists a constant $\delta > 0$ such that for all $t < T$, $\mathbb{E}[X_t - X_{t+1} \mid \mathcal{F}_t] \ge \delta$, then
\[
\mathbb{E}[T \mid X_0] \le \frac{X_0 - a}{\delta}.
\]
\end{theorem}



\begin{theorem}[Law of Total Expectation / Tower Property \textcolor{black}{\cite{M2017probability}}] \label{thm:tower-property}
Let $(\Omega, \mathcal{F}, \Pr)$ be a probability space, and let $\mathcal{F}_1$ and $\mathcal{F}_2$ be two sub-$\sigma$-algebras of $\mathcal{F}$ such that $\mathcal{F}_1 \subseteq \mathcal{F}_2$. For any random variable $X$ with finite expectation ($\mathbb{E}[|X|] < \infty$), it holds almost surely that:
\[
\mathbb{E}[\mathbb{E}[X \mid \mathcal{F}_2] \mid \mathcal{F}_1] = \mathbb{E}[X \mid \mathcal{F}_1].
\]
In the context of discrete-time stochastic processes with a natural filtration $(\mathcal{F}_t)_{t \ge 0}$, this implies that for any time steps $s \le t$, conditioning on the historical state $\mathcal{F}_s$ absorbs the intermediate conditioning on $\mathcal{F}_t$,
\[
\mathbb{E}[\mathbb{E}[X \mid \mathcal{F}_t] \mid \mathcal{F}_s] = \mathbb{E}[X \mid \mathcal{F}_s].
\]
\end{theorem}


\subsection{Notation}
Throughout this paper, the optimization process is modeled as a discrete-time stochastic process. We denote the current time step, which exactly corresponds to the number of fitness evaluations, by $t \in \mathbb{N}_0$, and the candidate solution maintained by the algorithm at step $t$ by $s^{(t)} \in \{0, 1\}^n$. The fitness value of the current solution is typically denoted by $i = \LO(s^{(t)})$. To rigorously ground our probabilistic arguments, we let $\mathcal{F}_t$ be the natural filtration representing the complete historical trajectory of the algorithmic state (including all past solutions, weights, and operator selections) up to step $t$. 

Regarding the operator selection mechanism, $w_j^{(t)}$ represents the current weight of heuristic $h_j$ (where $j \in \{1, 2\}$ respectively correspond to $\RLS_1$ and $\RLS_2$), and $p_j^{(t)} = w_j^{(t)} / (w_1^{(t)} + w_2^{(t)})$ defines the selection probability of that heuristic at step $t$. Furthermore, we define $P_j(i)$ as the {\it success} probability that applying heuristic $h_j$ to a solution with fitness $i$ yields a strict fitness improvement. Finally, $\Delta_j^{sel}$ denotes the conditional expected one-step change in the weight of heuristic $h_j$, given that it was selected for application at the current step. We say that an event occurs {\it asymptotically almost surely} (a.a.s.) if its probability is at least $1 - o(1)$.

\section{RLHH has Optimal Runtime for Leading Ones}


In this section, we rigorously prove that the RLHH algorithm achieves the optimal expected runtime for the \textsc{LeadingOnes} function. To understand the algorithmic dynamics, we first evaluate the state-dependent success probabilities of the two mutation operators at any given fitness level $i = \LO(s)$. 

For operator $\RLS_1$, a strict fitness improvement occurs if and only if the first $0$-bit is flipped. Thus, its success probability is uniformly bounded by $P_1(i) = 1/n$. For operator $\RLS_2$, a success requires flipping the first $0$-bit alongside any of the $n-i-1$ trailing bits. The exact probability of this joint event evaluates to:
$$ P_2(i) = \xmq{\frac{(n-i-1)}{\binom{n}{2}}}  = \frac{2(n - i - 1)}{n(n-1)}. $$

Comparing these transition probabilities reveals a strict phase transition in the optimal search strategy. Solving $P_2(i) > P_1(i)$ yields $i < n/2$. Consequently, $\RLS_2$ is preferable during the first half of the \xmq{optimization process}, whereas $\RLS_1$ becomes more effective once $i > n/2$. As established in the literature, the optimal strategy for an algorithm equipped with these two operators is to apply $\RLS_2$ exclusively for optimizing the first $n/2$ bits, and then permanently switch to $\RLS_1$ for the remainder of the optimization~\cite{LOW2020b}.

To demonstrate that the RLHH algorithm learns and executes this optimal operator sequence a.a.s., we divide our mathematical analysis into three subsequent subsections. First, we analyze Phase 1 ($i < n/2$), proving that the parameter adaptation mechanism rapidly establishes a strict dominance of $\RLS_2$ a.a.s. during the initial stage of optimization. Next, we examine Phase 2 ($i \geq n/2$), demonstrating that as the fitness crosses the theoretical threshold, the algorithm successfully reverses its weight distribution to maintain the absolute dominance of $\RLS_1$ a.a.s.. Finally, we synthesize the bounded times from these two distinct phases to compute the total expected optimization time, concluding that RLHH has the optimal expected runtime up to a lower-order $o(n^2)$ term.

\subsection{Dominance of $\RLS_2$ in Phase 1}
In this subsection, we analyze the runtime behavior of the RLHH algorithm on the \textsc{LeadingOnes} problem during Phase 1, defined as the interval where the fitness satisfies $0 \le \LO(s) < \frac{n}{2}$. In this region, the operator $\RLS_2$ possesses a higher success probability than the operator $\RLS_1$. The analysis demonstrates that the dynamic weight adaptation mechanism successfully identifies and exploits $\RLS_2$ with a selection probability of $p_2 = 1 - o(1)$, incurring an $o(n^2)$ time overhead \new{compared to the optimal expected runtime}.

The stochastic dynamics of Phase 1 are established through a sequence of supporting lemmata. First, Lemma~\ref{lem:drifts} derives the conditional 
drifts of the operator weights based on their state-dependent success probabilities, \new{showing that when the respective operators are selected $w_2$ grows asymptotically faster than $w_1$ in expectation}. \xu{Lemma~\ref{lem:no-rls1-dom} ensures that the selection probability $p_2^{(t)}$ remains bounded away from zero for Phase 1. Building on this, Lemma~\ref{lem:w2-reaches-max} proves that the weight $w_2$ rapidly reaches its upper limit $w_{\max}=\frac{n}{\log^{1+\gamma} n}$ within an $o(n^2)$ time overhead. Concurrently, Lemma~\ref{lem:w1-bound-mult} constructs a non-negative supermartingale to bound the multiplicative growth of $w_1$ during the climbing stage of $w_2$, ensuring $w_1$ remains sub-polynomial. Finally, Lemma~\ref{lem:maintenance} applies the law of total expectation, Doob's inequality and Freedman's inequality to show that the dominance of $\RLS_2$ is robustly maintained a.a.s. until the fitness reaches $n/2$.} 

\new{We put everything together in} Theorem~\ref{thm:phase1-time} to formally establish that RLHH \new{optimises the first $n/2$ leading ones in} expected time 
\new{$T_{\textsc{Opt}_{n/2}} + o(n^2)$, where $T_{\textsc{Opt}_{n/2}}$ is the exact expected hitting time for the pure $\RLS_2$ algorithm to traverse the same fitness distance.}

\new We 
start by analysing the expected single-step changes of the operator weights \new{conditional on the operators being selected for reproduction.}

\begin{lemma} \label{lem:drifts}
For the RLHH algorithm with reward parameter $\alpha = 1 + 1/\log n$ and penalty parameter $\beta = 1/n$ optimizing the \textsc{LeadingOnes} function, let $\Delta_1^{sel}$ and $\Delta_2^{sel}$ denote the expected one-step weight changes of the operators $\RLS_1$ and $\RLS_2$, conditioned on being selected. The weight drifts always satisfy
\[
    \Delta_1^{sel} = \frac{1}{n\log n} + \frac{1}{n^2}, \quad \text{and} \quad \Delta_2^{sel} = \mathcal{O}\!\left(\frac{1}{n}\right).
\]
Furthermore, for any fitness level $i \le n/2 - \epsilon n$ with any small constant $\epsilon > 0$, there exists a constant $c > 0$ such that the weight drift of $\RLS_2$ is lower-bounded by
\[
    \Delta_2^{sel} \ge \frac{c}{n}.
\]
\end{lemma}

\xu{By defining a composite process $U_t = 2w_2^{(t)} - w_1^{(t)}$ and  constructing an exponential potential function $Z_t = \exp(-\lambda U_t)$ that operates as a non-negative supermartingale, we rigorously bound the probability that $w_1$ ever exceeds twice the value of $w_2$ over a polynomial time horizon by applying Doob's maximal inequality (Theorem~\ref{thm:doob}) to $Z_t$.}

\begin{lemma}
\label{lem:no-rls1-dom}
For the RLHH algorithm optimizing the \textsc{LeadingOnes} function with initial weights $w_1^{(0)} = w_2^{(0)} = \log n$, $\alpha = 1 + 1/\log n$, $\beta = 1/n$, consider a polynomial time horizon $T = \operatorname{poly}(n)$. Assuming the fitness satisfies $\LO(s^{(t)}) \le n/2 - \epsilon n$ for a constant $\epsilon > 0$ for all $t \le T$, the probability that $w_1^{(t)} \ge 2 w_2^{(t)}$ occurs at any step $t \in [0, T]$ is bounded by
$$ \Pr\left( \exists t \in [0, T] : w_1^{(t)} \ge 2 w_2^{(t)} \right) = n^{-\Omega(1)}. $$
\end{lemma}

\new{To prove that the RLHH algorithm achieves the optimal Phase 1 runtime in Theorem~\ref{thm:phase1-time}, the selection probability of $\RLS_2$ must rapidly approach $1$. We first show that the weight $w_2$ climbs to its upper bound $w_{\max}$ in $o(n^2)$ steps \xu{with probability $1-o(1)$}. 
Within this time with overwhelming probability at most 
$\frac{n}{2} - \epsilon n$ leading ones are optimized for a constant $\epsilon > 0$. Lemma~\ref{lem:drifts}, thus ensures an  $\Omega(1/n)$ drift for $w_2$ during the climbing stage. 
}

\begin{lemma}
\label{lem:w2-reaches-max}
For the RLHH algorithm optimizing the \textsc{LeadingOnes} function with parameters $w_{\max} = \frac{n}{\log^{1+\gamma} n}$ for a constant $0<\gamma<1$, $\alpha = 1 + 1/\log n$, $\beta = 1/n$, with probability at least $1 - \mathcal{O}\!\left(\frac{1}{\log^{\gamma/2} n}\right)$, the weight $w_2$ reaches $w_{\max}$ within $t_{\max} = \frac{n^2}{\log^{1+\gamma/2} n}$ steps, and the fitness  satisfies $\LO(s^{(t)}) \le n/2 - \epsilon n$ during this entire climbing period.
\end{lemma}

While Lemma~\ref{lem:w2-reaches-max} proves that the weight $w_2$ rapidly reaches $w_{\max}$, establishing a selection probability of $p_2 = 1 - o(1)$ also requires that $w_1$ remains comparatively small 
during the climbing stage of $w_2$. Relying on the $o(n^2)$ time bound established in Lemma~\ref{lem:w2-reaches-max}, we 
analyze the multiplicative \new{increase} of $w_1$. 

We divide the climbing process of the weight of $\RLS_2$ from $w_2=\log n$ to $w_{\max}$ into $m$ phases of multiplicatively increasing length and for each phase we construct a non-negative supermartingale that allows us to prove using Freedman's inequality that with exponentially small probability in the length of the phase, the weight never drops below half the value it had at the beginning of the phase.
Then we construct another martingale for each phase to show that the probability that $w_2$ reaches the end of the phase in more than twice its expected time is also exponentially small. 
 Finally using the calculated bounds for $w_2$ we show
that $w_1$ remains  sub-polynomial i.e., much smaller than
$w_{\max}$.

\begin{lemma}
\label{lem:w1-bound-mult}
For the RLHH algorithm optimizing the \textsc{LeadingOnes} function with initial weights $w_1^{(0)} = w_2^{(0)} = \log n$, $\alpha = 1 + 1/\log n$, $\beta = 1/n$, let $\tau$ be the first time step where $w_2^{(\tau)}$ reaches $w_{\max} = \frac{n}{\log^{1+\gamma} n}$ for a constant $0<\gamma<1$. At time $\tau$, the weight $w_1$ is  bounded by $w_1^{(\tau)} \xmq{=} \ \mathcal{O}(\log^2 n)$ with probability at least $1 - \mathcal{O}\!\left(\frac{1}{\log n}\right)$.
\end{lemma}

To complete the analysis of Phase 1, we prove that once the weight $w_2$ reaches $w_{\max}$, it stably maintains its dominant position until the algorithm successfully reaches the fitness level $n/2$. Because the dynamics of $w_1$ and $w_2$ are coupled, we analyze them sequentially: we first unconditionally bound the maximum possible growth of $w_1$ assuming $w_2$ does not drop by more than half its value. Afterwards we use this established bound on $w_1$ to prove via Freedman's inequality that $w_2$ still does not drop by half its value  with high probability.

\begin{lemma}
\label{lem:maintenance}
For the RLHH algorithm optimizing the \textsc{LeadingOnes} function with parameters $w_{\max} = \frac{n}{\log^{1+\gamma} n}$ for a constant $0 < \gamma < 1$, $\alpha = 1 + 1/\log n$, $\beta = 1/n$, suppose $w_2$ has reached $w_{\max}$. With probability at least $1 - n^{-1/2+o(1)}$, the weights maintain the strict bounds $w_2^{(t)} \ge w_{\max}/2$ and $w_1^{(t)} \le \xmq{\sqrt{w_{\max}}}$ for all subsequent steps until Phase 1 concludes. Consequently, the selection probability robustly satisfies $p_2^{(t)} = 1 - o(1)$ throughout this entire period.
\end{lemma}

\new{We now prove the main result for Phase 1.}
\begin{theorem}
\label{thm:phase1-time}
For the RLHH algorithm optimizing the \textsc{LeadingOnes} function with parameters $w_1^{(0)} = w_2^{(0)} = \log n$, $w_{\max} = \frac{n}{\log^{1+\gamma} n}$ for a constant $0<\gamma<1$, $w_{\min}=1$, $\alpha = 1 + 1/\log n$, $\beta = 1/n$, let $T_{\text{phase1}}$ be the total number of fitness evaluations required to reach a fitness of \xu{at least} $n/2$ from a uniformly distributed random initial string. The expected runtime is strictly bounded by
$$ \mathbb{E}[T_{\text{phase1}}] = T_{\textsc{Opt}_{n/2}} + o(n^2), $$
where $T_{\textsc{Opt}_{n/2}}$ is the expected runtime for the theoretically optimal $\RLS_2$ algorithm to traverse the same fitness distance.
\end{theorem}

\subsection{Dominance of $\RLS_1$ in Phase 2}
\xu{In this section, we bound the expected runtime of RLHH during Phase 2. For the purpose of analysis, we reset the time counter to $t=0$ at the transition point and pessimistically consider a worst-case initial weight distribution, where $w_1^{(0)} = w_{\min}$ and $w_2^{(0)} = w_{\max}$.}

\xu{First, Lemma~\ref{lem:buffer-zone} uses the Artificial Fitness Level Method~\cite{OlivetoHandbook} to establish that the algorithm quickly reaches a sufficiently high fitness level where $\RLS_2$ starts to lose its advantage quickly. Then, Lemma~\ref{lem:weight-boundary} proves that once the algorithm reaches this high-fitness region, at least one of the weights will rapidly hit a boundary. Once at least one boundary is hit, Lemma~\ref{lem:maintenance} guarantees that $\RLS_1$ will rapidly become dominant and maintain its dominance until the end of the optimization process.}

\xu{We put these lemmata together in Theorem~\ref{thm:phase2-time} to establish that RLHH optimizes the remaining fitness levels from $n/2$ to $n$ in expected time $T_{\textsc{Opt}_{n/2 \to n}} + o(n^2)$, where $T_{\textsc{Opt}_{n/2 \to n}}$ is the exact expected hitting time for the pure $\RLS_1$ algorithm to traverse this fitness distance.}

\begin{lemma}
\label{lem:buffer-zone}
For the RLHH algorithm optimizing the \textsc{LeadingOnes} function, assume the algorithm has reached the beginning of Phase 2 with fitness $\LO(s^{(0)}) = n/2$.
\xu{The expected time to reach a fitness of at least $n/2 + n/\log n$ is bounded by $\mathbb{E}[T_{\text{buf}}] \le \frac{2n^2}{\log n}$.}
\end{lemma}

We now bound the expected time required for either weight to reach its boundary.
In particular\xmq{,} we define a potential function that captures the distance of the two weights from the respective boundaries.
Then we apply the additive drift theorem to calculate the expected time for the potential function to reach zero which will imply that one of the two boundaries has been reached.
\begin{lemma}
\label{lem:weight-boundary}
For the RLHH algorithm optimizing the \textsc{LeadingOnes} function with parameters $w_{\max} = \frac{n}{\log^{1+\gamma} n}$ for a constant $0<\gamma<1$, $w_{\min} = 1$, $\alpha = 1 + 1/\log n$, $\beta = 1/n$, assume the algorithm has reached the beginning of Phase 2 with fitness $\LO(s^{(0)}) = n/2$.
Let $\tau$ be the first time step where either the condition $w_1^{(t)} = w_{\max}$ or $w_2^{(t)} = w_{\min}$ is triggered.
\xu{The expected hitting time is bounded by $\mathbb{E}[\tau] = \mathcal{O}\!\left(\frac{n^2}{\log^\gamma n}\right)$.}
\end{lemma}

\xu{In the following lemma, we prove that once either weight hits its respective boundary, the system rapidly stabilizes into a new regime that strictly favors $\RLS_1$ (e.g., $p_1 = 1 - o(1)$). }
We split the proof into two cases.

For the case where $w_1$ has hit the upper boundary, we first use Freedman's inequality to show that $w_1$ never drops by more than half its value. Using this bound on $w_1$, we calculate the expected multiplicative decrease of $w_2$ assuming pessimistically that at the beginning of the phase its value is $w_{\max}$.
Then the law of total expectation and a Markov inequality allow us to show that  a.a.s. $w_2$ drops down to at least $o(w_{\max})$ in $o(n^2)$ steps. Finally, we apply Doob's inequality to show that
$w_2$ remains $o(w_{\max})$ until the global optimum is found.

For the second case where $w_2$ has hit the lower boundary, we first use \xmq{Doob's inequality} to show that $w_2$ never increases above $\log^2 n$. \xmq{Then} we use this bound on the weight to calculate the additive drift of $w_1$ which allows us to derive the expected time for $w_1$ to reach a high enough value to guarantee a selection probability of $p_1= 1-o(1)$.
A Markov inequality and a Freedman\xmq{'s} inequality show that $p_1=1-o(1)$ is sustained until the end of the phase.

\xu{\begin{lemma}
\label{lem:new-regime}
For the RLHH algorithm optimizing the \textsc{LeadingOnes} function with parameters $w_{\max} = \frac{n}{\log^{1+\gamma} n}$ for a constant $0<\gamma<1$, $w_{\min} = 1$, $\alpha = 1 + 1/\log n$, $\beta = 1/n$, suppose at time $\tau$ the weights satisfy either $w_1^{(\tau)} = w_{\max}$ or $w_2^{(\tau)} = w_{\min}$. With probability at least $1 - \mathcal{O}\!\left(\frac{1}{\log^{2} n}\right)$, within an additional $t_{\text{stab}} = \frac{n^2}{\log^{\gamma/2} n}$ steps, the algorithm establishes a dominance regime where the selection probability of $\RLS_1$ satisfies $p_1^{(t)} = 1 - o(1)$, and this regime is maintained until Phase 2 concludes.
\end{lemma}}

\xu{We now put all the lemmata together to prove the main result for Phase 2.}

\begin{theorem}
\label{thm:phase2-time}
For the RLHH algorithm optimizing the \textsc{LeadingOnes} function with parameters $w_1^{(0)} = w_2^{(0)} = \log n$, $w_{\max} = \frac{n}{\log^{1+\gamma} n}$ for a constant $0<\gamma<1$, $w_{\min} = 1$, $\alpha = 1 + 1/\log n$, $\beta = 1/n$, let $T_{\text{phase2}}$ be the total number of fitness evaluations required to reach the global optimum $\LO(s) = n$ starting from a fitness of $n/2$. The expected runtime is strictly bounded by
$$ \mathbb{E}[T_{\text{phase2}}] = T_{\textsc{Opt}_{n/2 \to n}} + o(n^2), $$
where $T_{\textsc{Opt}_{n/2 \to n}}$ is the exact expected hitting time for the theoretically optimal pure $\RLS_1$ algorithm to traverse the second half of the search space.
\end{theorem}

\subsection{Total Expected Runtime}

In this subsection, we summarize the runtime analyses of Phase 1 and Phase 2 to establish the overall expected runtime of the RLHH algorithm. We show that the weight adaptation successfully matches the performance of the theoretically optimal operator sequence, accumulating only a sub-quadratic time overhead across the entire optimization process.

\begin{theorem}
\label{thm:total-runtime}
For the RLHH algorithm optimizing the \textsc{LeadingOnes} function with parameters $w_1^{(0)} = w_2^{(0)} = \log n$, $w_{\max} = \frac{n}{\log^{1+\gamma} n}$ for a constant $0<\gamma<1$, $w_{\min} = 1$, $\alpha = 1 + 1/\log n$, $\beta = 1/n$, let $T$ be the total number of fitness evaluations required to reach the global optimum $\LO(s) = n$ from a uniformly distributed random initial string. The expected runtime is strictly bounded by
$$ \mathbb{E}[T] = T_{\textsc{Opt}_{n/2}} + T_{\textsc{Opt}_{n/2 \to n}} + o(n^2) = \frac{1+\ln 2}{4}n^2 + o(n^2). $$
\end{theorem}





\section{Experimental Evaluation}

\begin{figure}[htbp]
    \centering
    \includegraphics[width=0.8\textwidth]{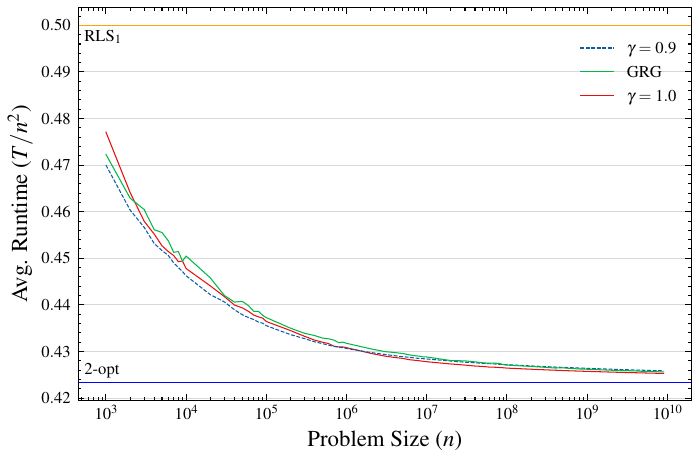}
    \caption{Average runtime normalized by $n^2$ for the RLHH algorithm ($\gamma \in \{0.9, 1.0\}$) compared against GRG, $\RLS_1$, and the theoretical 2-opt optimal strategy on the \textsc{LeadingOnes} function.}
    \label{fig:runtime_comparison}
\end{figure}

In this section, we empirically evaluate the performance of the RLHH algorithm on the \textsc{LeadingOnes} function for realistic problem sizes and compare it against the Greedy Random Gradient (GRG) HH which also has optimal expected runtime up to lower order terms~\cite{LOW2020b}. We equip both HHs with the heuristic set $H=\{\RLS_1,\RLS_2\}$.


We simulate the algorithms across a wide range of problem sizes, from $n = 10^3$ to $n = 9\cdot 10^{9}$ and for each problem size we take the average number of function evaluations over 700 runs. 
To allow to experiment with large problem sizes we apply the same inverse transform sampling method previously used for GRG in~\cite{LOW2020b} and sample the expected waiting times to make an improvement of exactly $d$ bits, for all possible values of $d$ and for each operator instead of executing the operator itself.

 For the RLHH algorithm, we utilize the theoretically derived parameter settings: 
 $\alpha = 1 + 1/\log n$, 
 $\beta = 1/n$, initial weights $w_1^{(0)} = w_2^{(0)} = \log n$, and the upper bound $w_{\max} = n/\log^{1+\gamma} n$. 
Our theoretical analysis indicates that $w_{\max}$ should be as small as possible, thus $\gamma$ as close \xmq{to} $1$ as possible. This respectively decreases the runtimes derived in \xmq{Lemmata} \ref{lem:w2-reaches-max}, \ref{lem:weight-boundary} and \ref{lem:new-regime} and the relevant failure probabilities. Indeed, preliminary tuning confirms this for these problem sizes. Thus, we test the algorithm with
 $\gamma = 0.9$ and $\gamma = 1.0$. For the GRG we set $\tau=0.6 n \ln n$ which provided the best performance in Lissovoi et al.~\cite{LOW2020b}. 


Figure~\ref{fig:runtime_comparison} displays the average runtime normalized by $n^2$ (i.e., the leading constants) versus the problem size.
We observe 
that the runtimes of all three algorithms
exhibit a clear, monotonic downward trend, steadily approaching the best possible expected runtime using the two operators of $\frac{1+\ln 2}{4}n^2 \approx 0.423n^2$ (i.e., 2-opt indicated in blue).
The theoretical performance of the best randomised local search heuristic for the problem (i.e. $\RLS_1$ with expected runtime $0.5 n^2$) is depicted in yellow. 

The configuration $\gamma = 1.0$ has the best average runtime, consistently outperforming GRG for all problem sizes greater than $n=2\cdot10^3$.
For the largest problem size, $n=9\cdot10^9$, the two algorithms have average runtimes of $0.425n^2$  
and $0.426n^2$ 
slightly larger than the best possible $0.423n^2$.
While our formal proof conservatively requires $\gamma < 1$ to mathematically guarantee that $w_1 = o(w_{\max})$ at the conclusion of Phase 1, the practical growth rate of the weight (bounded by $e^{c \log^\gamma n}$) is sufficiently slow. Consequently, the experiments indicate that setting $\gamma = 1.0$ provides an excellent balance between rapid weight adaptation and the stability of the dominance regimes in practice. On the other hand, the configuration with $\gamma = 0.9$ consistently outperforms the other two algorithms for smaller problem sizes and eventually becomes only slightly worse than GRG.

The only other reinforcement learning heuristic for which a theoretical runtime analysis is available is the theory-driven designed one introduced by Doerr et al.~\cite{DoerrEtAl2016}.
The authors proved optimal performance for \textsc{OneMax} and provided experiments for $\LeadingOnes$. The best average runtime they reported with their chosen parameters is $0.45n^2$ for $n=10^4$ which is beaten by all three algorithms in Figure~\ref{fig:runtime_comparison}.


\section{Conclusion}

In this paper, we presented a rigorous runtime analysis of the Reinforcement Learning Hyper-heuristic (RLHH) for the standard \textsc{LeadingOnes} benchmark function.
This is the first time a \xmq{rigorous positive result is presented regarding a} hyper-heuristic using a reinforcement learning mechanism from the literature. 
The only previous theoretical analysis had considered the RLHH equipped with only one useful operator having positive improvement probability~\cite{AL2016}.   
The authors proved that for a wide range of parameter values (i.e., $\alpha \leq \beta$) RLHH is ineffective for \textsc{LeadingOnes} by showing that its behaviour is similar to that of a Simple Random HH that at each step selects an operator uniformly at random (i.e. the algorithm does not learn to identify the only good operator).

Our analysis proves that RLHH equipped with a low-level heuristic set $H=\{\RLS_1,\RLS_2\}$ and appropriate parameter values adapts the weights optimally throughout the process, thus optimises \textsc{LeadingOnes} in an expected runtime of $\frac{1+\ln 2}{4}n^2 + o(n^2)$ perfectly matching, up to the leading constant, the best possible offline-designed sequence of these operators.
The proof of the result relies on tracking the stochastic weight dynamics via various martingale processes and drift analyses.
An empirical analysis shows that the algorithm is effective for realistic problem sizes even outperforming the Generalised Random Gradient (GRG) HH previously shown to also have optimal expected runtime.



Future research should extend the analysis to larger heuristic set sizes and to wider classes of functions. 
Furthermore the comparative performance using alternative update and selection rules from the literature should also be analysed such as multiplicative updates or the {\it max choice} selection rule.  
\bibliographystyle{splncs04}
\bibliography{references}
%
%
%
%
%
%




\end{document}